\documentclass[letterpaper, 10 pt, conference]{ieeeconf}  

\IEEEoverridecommandlockouts                              

\usepackage{epsfig} 
\usepackage{subfigure}
\usepackage{amsmath} 
\usepackage{amssymb}  
\usepackage{makecell}
\usepackage{multirow}
\usepackage{float}
\usepackage[colorlinks=true,linkcolor=blue,citecolor=green,urlcolor=blue,]{hyperref}

\title{\LARGE \bf
PLV-IEKF: Consistent Visual-Inertial Odometry using Points, Lines, and Vanishing Points
}

\author{Tong Hua$^{1}$, Tao Li$^{1}$, Liang Pang$^{2}$, Guoqing Liu$^{1}$, Wencheng Xuanyuan$^{2}$, Chang Shu$^{2}$ and Ling Pei$^{*,1}$
\thanks{*This work was supported in part by the National Nature Science Foundation of China (NSFC) under Grant 62273229, and in part by Shanghai Slamtec Co., Ltd. and completed in cooperation with Shanghai Slamtec Co., Ltd.}
\thanks{$^{1}$ Tong Hua, Tao Li, Ling Pei and Guoqing Liu are with Shanghai Key Laboratory of Navigation and Location Based Services, Shanghai Jiao Tong University. $^*$ Corresponding Author {\tt\small ling.pei@sjtu.edu.cn}}%
\thanks{$^{2}$ Liang Pang, Wencheng Xuanyuan and Chang Shu are with SLAMTEC company.}%
}

\begin{document}

\maketitle
\thispagestyle{empty}
\pagestyle{empty}

\begin{abstract}
In this paper, we propose an Invariant Extended Kalman Filter (IEKF) based Visual-Inertial Odometry (VIO) using multiple features in man-made environments. Conventional EKF-based VIO usually suffers from system inconsistency and angular drift that naturally occurs in feature-based methods. However, in man-made environments, notable structural regularities, such as lines and vanishing points, offer valuable cues for localization. To exploit these structural features effectively and maintain system consistency, we design a right invariant filter-based VIO scheme incorporating point, line, and vanishing point features. We demonstrate that the conventional additive error definition for point features can also preserve system consistency like the invariant error definition by proving a mathematically equivalent measurement model. And a similar conclusion is established for line features. Additionally, we conduct an invariant filter-based observability analysis proving that vanishing point measurement maintains unobservable directions naturally. Both simulation and real-world tests are conducted to validate our methods' pose accuracy and consistency. The experimental results validate the competitive performance of our method, highlighting its ability to deliver accurate and consistent pose estimation in man-made environments.
\end{abstract}

\section{INTRODUCTION}

Accurate pose estimation is a fundamental issue for robot localization, and a range of approaches have been developed in the previous literature which investigates different sensor fusion methods like GNSS \cite{cao2022gvins, lee2020intermittent, li2022p, hua2023m2c} and LiDAR \cite{zhang2014loam,li2023p}. As an attractive candidate, Visual-Inertial Odometry (VIO) has been widely studied for its high accuracy, low cost, and lightweight. Moreover, Extended Kalman Filter (EKF) based VIOs like Multi-State Constraint Kalman Filter (MSCKF) \cite{mourikis2007multi} have been favored by many researchers for its great advantage in high computational efficiency.

Conventional MCSKF utilizes point features to constrain the state. However, it may produce a large pose drift in a textureless environment or illumination changing scenes due to the limited number of features in the scene \cite{zhou2015structslam}. As an alternative solution, line features may provide additional structural information in challenging environments \cite{he2018pl}. Furthermore, vanishing points (VP) in man-made scenarios can make line feature fully observable \cite{lim2022uv}. Therefore, these structural features can be fully leveraged in VIO to obtain a more accurate pose estimation \cite{zou2019structvio, heo2018consistent, lee2021plf}. 

Standard EKF frameworks, regardless of whether they use points or lines, have been proven to be inconsistent due to spurious information along the unobservable direction \cite{heo2018consistent}. To address this issue, various algorithms have been proposed. In recent years, the Invariant EKF (IEKF) has been successfully applied in robotic localization, particularly in filter-based VIO \cite{wu2017invariant, brossard2018invariant,zhang2023toward,jung2022photometric}. The IEKF model defines an alternative nonlinear error for estimated poses and landmarks, automatically ensuring the appropriate dimension of the unobservable subspace \cite{barrau2015ekf}. However, in the case of MSCKF, landmarks are not included in the state. This leads to two different error definitions for landmarks: the nonlinear error definition following IEKF and the additive error definition following the conventional MSCKF. The relationship between these two definitions and their impact on filter consistency is our motivation to investigate.

Our main contributions are summarized as follows:
\begin{itemize}
\item We present a right invariant filter-based VIO leveraging points, lines, and vanishing points, which improves both the pose consistency and accuracy. 
\item In the filter design, we prove an equivalent measurement model for point features, demonstrating that these features do not change the filter consistency even in the conventional error representation when they are decoupled from the state vector. And a similar conclusion can be given for line features.
\item An observability analysis is given to demonstrate that our method maintains the system’s unobservable subspace naturally and enhances the observability of line features.
\end{itemize}

\section{RELATED WORKS}
\subsection{IEKF Applications for VIO}
IEKF \cite{bonnabel2007left} is proposed to address the filter inconsistency problem, which has been widely researched in a range of literature \cite{barrau2015ekf, bailey2006consistency, huang2007convergence, huang2008analysis,li2013high, huang2010observability,hua2023piekf}. The IEKF model has been further applied to MSCKF \cite{heo2018consistent, wu2017invariant, 9844243} and UKF \cite{brossard2018invariant}. Previous works \cite{wu2017invariant}, \cite{brossard2018unscented} have coupled the landmark uncertainty with the camera pose, while other works adopt the simple addictive error \cite{9844243}. Though both strategies can maintain the consistency, the relationship between two kinds of error definition has not been investigated.

\subsection{Point-Line-VIO}
A couple of works have combined point and line features. Among optimization-based methods, \cite{he2018pl} jointly minimizes the IMU pre-integration constraints together with the point and line re-projection errors. \cite{fu2020pl} further modifies the LSD \cite{von2008lsd} algorithm to have a better real-time performance based on \cite{he2018pl}. By utilizing the structural regularity, Lee et al. introduce parallel lines' constraints into point-line-based VIO and present a novel structural line triangulation method making full use of the prior information \cite{lee2021plf}. Among filter-based methods, the issue of point-line VIO inconsistency is addressed and improved by observability-constrained techniques \cite{kottas2013efficient} and the IEKF framework \cite{heo2018consistent}. \cite{yu2015vision} extends a new line parameterization for processing the line observations in a rolling shutter camera setting. Zheng et al. present Trifo-VIO which incorporates point and line feature measurements and formulate loop closure as EKF updates \cite{zheng2018trifo}. \cite{yang2019visual} investigates two line triangulation methods and reveals three degenerate motions which cause triangulation failures.

\subsection{Vanishing Point aided SLAM/VIO}
In man-made environments, geometric information like vanishing points can also boost the robustness of VIO and Simultaneous Localization and Mapping (SLAM) \cite{pumarola2017pl, gomez2019pl, li2018monocular, georgis2022vp}. Zhang et al. provide a solution to the loop closure problem leveraging vanishing points in line-based SLAM \cite{zhang2012loop}. \cite{camposeco2015using} proposes an efficient line classification method for detecting vanishing points and keeps the vanishing points in the VIO state for a long tracking. Xu et al. design a point-line-based VIO system where vanishing points are used to recognize and classify structural lines \cite{xu2022leveraging}. In \cite{lim2022uv}, the vanishing point cost function is added into the optimization-based VIO without the constraints like the Manhattan world assumption \cite{coughlan2000manhattan}. However, these works have not focused on consistency about vanishing point measurements.

\section{PRELIMINARY KNOWLEDGE}
\subsection{IEKF Model}
In the visual-inertial navigation system, the defined state is given by:
\begin{equation}
\mathbf{X} = (
\mathbf{R}, \mathbf{v}, \mathbf{p}, \mathbf{b}_g,\mathbf{b}_{a}, \mathbf{p}_f)\label{eq26}
\end{equation}
where $\mathbf p_f$ is the landmark. Unlike the conventional error in standard EKF, the right invariant filter employs a new uncertainty representation:
\begin{equation}
    \begin{aligned}
    \mathbf{X} =& (exp(\mathbf{\xi}_\theta)\hat{\mathbf{R}}, exp(\mathbf{\xi}_\theta)\hat{\mathbf{v}}+\mathbf{J}_l(\mathbf{\xi}_\theta)\mathbf{\xi}_v, exp(\mathbf{\xi}_\theta)\hat{\mathbf{p}}+\mathbf{J}_l(\mathbf{\xi}_\theta)\mathbf{\xi}_p,\\
    &\, \, \mathbf{\hat{b}}_g + \mathbf{\xi}_{b_g}, \mathbf{\hat{b}}_a + \mathbf{\xi}_{b_a}, exp(\mathbf{\xi}_\theta)\hat{\mathbf p}_f + \mathbf{J}_l(\mathbf{\xi}_\theta)\mathbf{\xi}_{p_f}) 
    \end{aligned}\label{eq5}
\end{equation}
where $\mathbf{J}_l$ is the left Jacobian for $SO3$ Lie group, and $\xi = \begin{bmatrix}
\xi_{\theta}^T & \xi_v^T & \xi_p^T & \xi_{b_g}^T & \xi_{b_a}^T & \xi_{p_f}^T
\end{bmatrix}^T$ is the state error vector. Note that apart from robot velocity and position, the landmark's uncertainty is also coupled with the rotation state. This nonlinear error allows the system to obtain a better performance on consistency.
\subsection{Line Representation}
There exist many representation models for a 3D line, and two representations are focused on in our paper. One is the conventional Plucker coordinate $\mathcal{L} = \begin{bmatrix} \mathbf{n}^T & \mathbf{d}^T \end{bmatrix}^T \in \mathbb{R}^6$, where $\mathbf n \in \mathbb{R}^3$ is the normal vector of the plane determined by the line and the coordinate origin, and $\mathbf d \in \mathbb{R}^3$ is the line direction vector \cite{he2018pl}. However, this representation has redundant parameters since the degrees of freedom for a line should be four, so the minimal representation $\mathbf o = \begin{bmatrix} \mathbf \psi^T & \phi \end{bmatrix}^T$ is more suitable in line optimization:
\begin{equation}
\begin{aligned}
    &\mathbf \psi = log(\mathbf U), \mathbf U =
\left[\frac{\mathbf{n}}{\|\mathbf{n}\|} \quad \frac{\mathbf{d}}{\|\mathbf{d}\|} \quad \frac{\lfloor \mathbf{n} \rfloor \mathbf{d}}{\|\lfloor \mathbf{n} \rfloor \mathbf{d}\|}\right]\\
    &\phi = arcsin(\frac{||\mathbf d||}{\sqrt{||\mathcal{L}||^2}}), \mathbf W = \begin{bmatrix}
        cos(\phi) & -sin(\phi) \\
        sin(\phi) & cos(\phi)
    \end{bmatrix}
\end{aligned} \label{eq27}
\end{equation}
where $\lfloor\cdot \rfloor$ denotes the skew symmetric matrix.

\section{FILTER DESCRIPTION}
\begin{figure}
    \centering
    \includegraphics[scale=0.45]{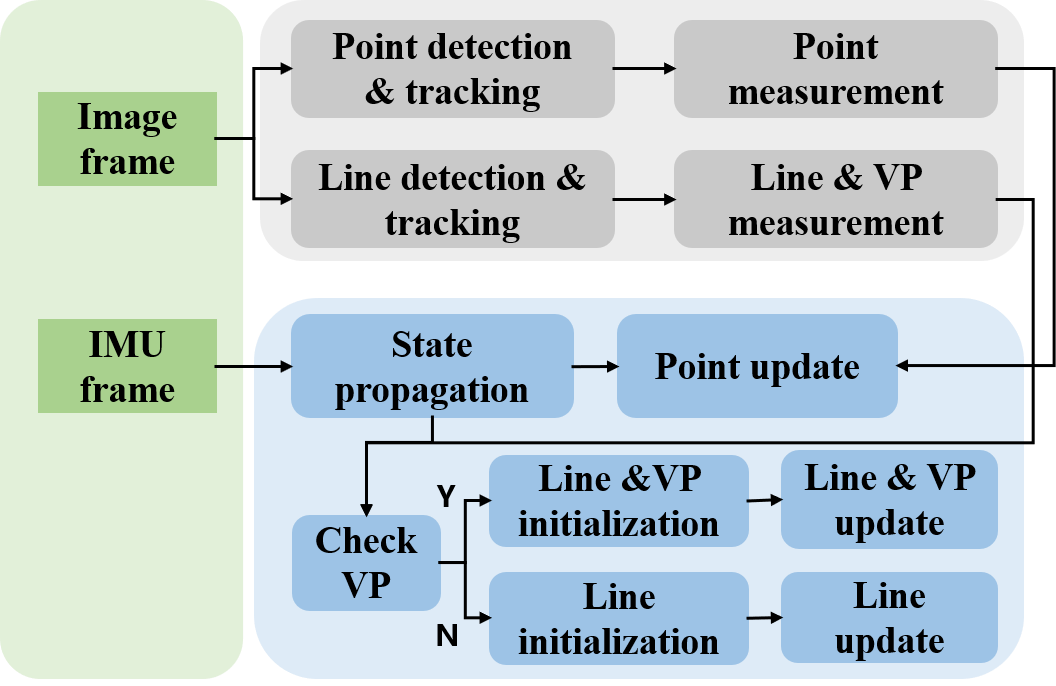}
    \caption{Overview of PLV-IEKF system.}
    \label{fig:overview}
\end{figure}
In the following sections, we denote the estimated state as $\hat{\mathbf x}$, and the error state as $\Tilde{\mathbf x}$. We build our system based on IEKF framework, as shown in Fig. \ref{fig:overview}. Specifically, we define the system state as:
\begin{align}
\mathbf{X} &= (
\mathbf{X}_I, \mathbf{X}_{C_1}, ... , \mathbf{X}_{C_n}) \label{eq1}\\
\mathbf{X}_I &= (
{ }_{I}^{G}\mathbf{R}, {}^G\mathbf{v}_I, { }^{G}\mathbf{p}_I, \mathbf{b}_g,\mathbf{b}_{a})\label{eq2}\\
\mathbf{X}_C &= ({}^G_C \mathbf{R}, {}^G \mathbf{p}_C)\label{eq3}
\end{align}
where $n$ is the sliding window length. $\mathbf X_I$ and $\mathbf X_C$ represents the IMU state and camera pose respectively, ${}^G_I\mathbf R$ is the rotation from the IMU frame to the global frame, ${}^G\mathbf v_I$ and ${}^G\mathbf p_I$ are the global velocity and position of IMU, respectively. $\mathbf b_g$ and $\mathbf b_a$ are the gyroscope and acceleromter biases, respectively. Note that we do not incorporate landmarks in $\mathbf X_I$, which is different from the conventional IEKF model in (\ref{eq26}). In other words, landmarks are decoupled from the state.

\subsection{IMU Kinematic Model}
The IMU continuous kinematic model is given by:
\begin{equation}
\begin{aligned}
    {}^G_I\dot{\mathbf{R}} &= {}^G_I\mathbf{R}\lfloor {}^I\pmb{\omega}-\mathbf{b}_g-\mathbf{n}_g\rfloor \\
    {}^G\dot{\mathbf{v}}_I &= {}^G_I\mathbf{R}({}^I\mathbf{a}-\mathbf{b}_a-\mathbf{n}_a+{}^I\mathbf{g}), {}^G\dot{\mathbf{p}}_I = {}^G\mathbf{v}_I\\
    \dot{\mathbf{b}}_g &= \mathbf{n}_{\omega g}, \dot{\mathbf{b}}_a = \mathbf{n}_{\omega a}
    \label{eq4}
\end{aligned}
\end{equation}
where ${}^I\pmb{\omega}$ and ${}^I\mathbf{a}$ are the angular velocity and linear acceleration velocity, respectively. $\mathbf{n}_{wg}$ and $\mathbf{n}_{wa}$ are Gaussian noises. After linearization, the following propagation equation is obtained:
\begin{equation}
    \dot{\Tilde{\mathbf{X}}}_I = \mathbf{F}\Tilde{\mathbf{X}}_I + \mathbf{G}\mathbf{n}_I \label{eq22}
\end{equation}
where $\mathbf{n}_I=\begin{bmatrix}\mathbf{n}_g^T & \mathbf{n}_{wg}^T & \mathbf{n}_a^T & \mathbf{n}_{wa}^T\end{bmatrix}^T$ is the process noise. $\mathbf{F}$ is the continuous state transition matrix and $\mathbf{G}$ is the input noise Jacobian.
\subsection{Point Measurement Model}

When the position of the point feature is initialized, the normalized coordinate $\mathbf z$ of the feature point ${}^G\mathbf{p}_f$ is obtained:
\begin{equation}
    \mathbf{z} = \pi({}^C\mathbf{p}_f) = \frac{1}{{}^C\mathbf{z}_f}\begin{bmatrix}
    {}^C\mathbf{x}_f \\ {}^C\mathbf{y}_f
    \end{bmatrix} = h(\mathbf{X},{}^G\mathbf{p}_f) \label{eq6}
\end{equation}
where $\pi$ is the projection function. The corresponding error vector of $\mathbf{X}$ is denoted as $\mathbf{\xi} = \begin{bmatrix}\mathbf{\xi}_I^T & \mathbf{\xi}_{C1}^T & ... & \mathbf{\xi}_{Cn}^T\end{bmatrix}^T \in \mathbb{R}^{15+6n}$. Therefore, the linearized measurement model can be obtained as below:
\begin{align}
    &\Tilde{\mathbf{z}} = \mathbf{H}_X\Tilde{\mathbf X} + \mathbf{H}_f{}^G\Tilde{\mathbf{p}}_f \label{eq7}
\end{align}
where Jacobians $\mathbf{H}_x$ and $\mathbf{H}_f$ are given by (\ref{eq:appendix}) in Appendix. Genrally, ${}^G\Tilde{\mathbf p}_f$ follows an invariant error definition like (\ref{eq5}):
\begin{equation}
    {}^G\mathbf p_f = exp(\xi_{\theta}^i) {}^G\hat{\mathbf{p}}_f + \mathbf{J}_l(\xi_{\theta}^i){}^G\Tilde {\mathbf{p}}_f \label{eq12}
\end{equation}
where $\xi_{\theta}^i$ is the orientation error for $\mathbf X_{C_i}$, and $C_i$ is the camera pose that captures the landmark first.
However, a proposition is given to reveal the equivalent relationship between the right invariant form and the conventional error form, i.e. ${}^G\Tilde{\mathbf{p}}_f = {}^G\mathbf{p}_f - {}^G\hat{\mathbf{p}}_f$.

$\mathbf{Proposition}$ $\mathbf{1.}$ Given the same estimated landmark, the measurement model for points is equivalent for the invariant error and additive error.

$\textit{Proof}$: See Appendix.

$\textit{Remark}$: The equivalent measurement model indicates the same consistency property. However, the additive error definition is more convenient to implement.
\subsection{Line Measurement Model}
\begin{figure}
    \centering
    \includegraphics[scale = 0.4]{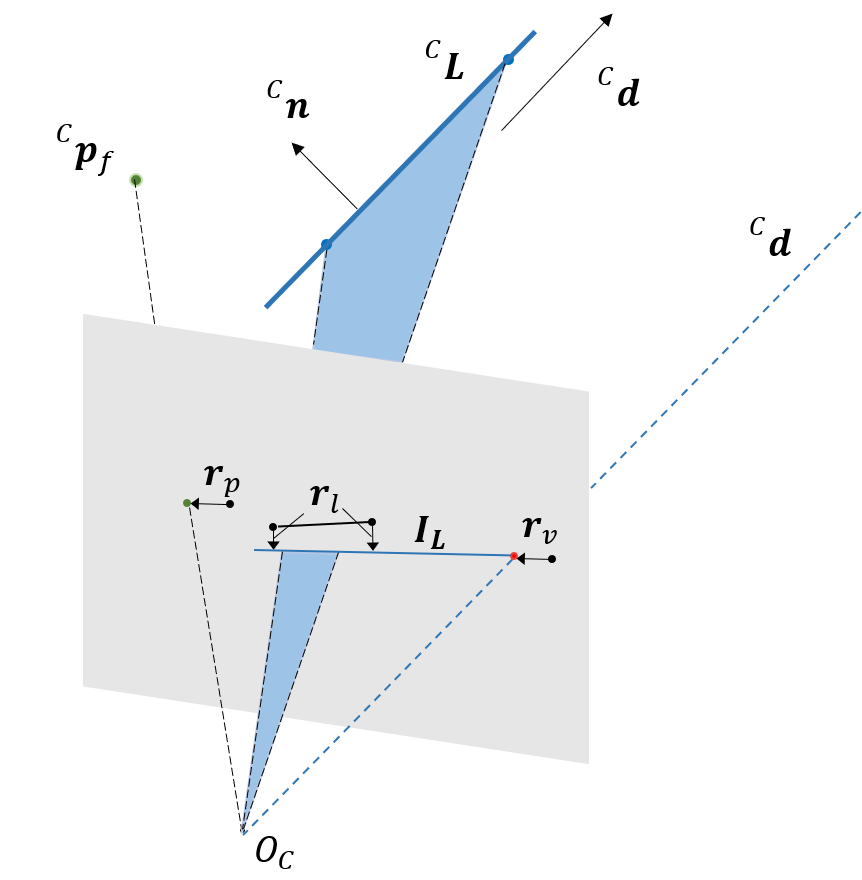}
    \caption{Illustration of point, line, and vanishing point residual. Three features are all projected to the normalized plane, and the black dots represent the measurements.}
    \label{fig:plv_residual}
\end{figure}
As shown in Fig. \ref{fig:plv_residual}, $\mathbf I_L$ represents the projection of a straight line $\mathcal{L}$ on the normalized image plane, so the error term is defined as the distance between the endpoints on the line in an image and the estimated 2D line:
\begin{equation}
    \Tilde{\mathbf z}_L = 
    \begin{bmatrix}
        z_{L,1} \\
        z_{L,2}
    \end{bmatrix} =
    \begin{bmatrix}
        \frac{\mathbf p_s^T\mathbf l}{\sqrt{l_1^2 + l_2^2}} \\
        \frac{\mathbf p_e^T\mathbf l}{\sqrt{l_1^2 + l_2^2}}
    \end{bmatrix} \label{eq10}
\end{equation}
where $\mathbf l = \begin{bmatrix} l_1 & l_2 & l_3\end{bmatrix}^T$ is the re-projected line on the normalized plane, $\mathbf p_s = [u_s \quad v_s \quad 1]^T$ and $\mathbf p_e = [u_e \quad v_e \quad 1]^T$ are the endpoint measurements of the line on the normalized plane.

The line error for the orthonormal representation is classified into globally-defined error and locally-defined error as below:
\begin{equation}
\begin{aligned}
    G: \mathbf{U} = exp(\delta \psi)\mathbf{\hat{U}}, \mathbf{W} = exp(\delta \phi)\mathbf{\mathbf{\hat W}} \\
    L: \mathbf{U} = \mathbf{\hat{U}}exp(\delta \psi), \mathbf{W} = \mathbf{\mathbf{\hat W}}exp(\delta \phi)
\end{aligned}\label{eq31}
\end{equation}
Thus, a conclusion on these two error forms can be given:

$\mathbf{Proposition}$ $\mathbf{2.}$ Given the same estimated line, the measurement models for lines are equivalent for the globally-defined error and locally-defined error.

$\textit{Proof}$: The proof is similar to that of $\mathbf{Proposition}$ 1.

$\textit{Remark}$: Conventional line-based VIO usually chooses the locally-defined error \cite{he2018pl} while IEKF-based VIO \cite{heo2018consistent} may choose the global one. Since the line feature is not incorporated in the state variables, both errors are acceptable in the algorithm implementation. 

Therefore, the linearized measurement model following a globally-defined error is given by:
\begin{equation}
    \begin{aligned}
    &\Tilde{\mathbf{z}}_L = \mathbf{H}_X\Tilde{\mathbf X}_C + \mathbf{H}_L\Tilde{\mathbf{o}} \label{eq29} \\
    &\mathbf{H}_X = \mathbf{J}_L{}^G_C\mathbf R^T
    \begin{bmatrix} (\lfloor {}^G\mathbf{n}\rfloor-\lfloor {}^G\mathbf{p}_{C} \rfloor \lfloor {}^G\mathbf{d} \rfloor) & \lfloor {}^G\mathbf{d} \rfloor \end{bmatrix} \\
    &\mathbf{H}_L = -\mathbf J_L{}^G_C\mathbf{R}^T\left[\left(\lfloor {}^G\mathbf{n}\rfloor-\lfloor {}^G\mathbf{p}_C \rfloor \lfloor {}^G\mathbf{d} \rfloor \right) \quad \mathbf{H}_{\phi} \right ] \\
    &\mathbf{J}_L = \begin{bmatrix}
    \frac{u_s}{l_{1,2}}-\frac{l_1 z_{L,1}}{l_{1,2}^{3}} & \frac{v_s}{l_{1,2}}-\frac{l_2 z_{L,1}}{l_{1,2}^{3}} & \frac{1}{l_{1,2}} \\
    \frac{u_e}{l_{1,2}}-\frac{l_1 z_{L,2}}{l_{1,2}^{3}} & \frac{v_e}{l_{1,2}}-\frac{l_2 z_{L,2}}{l_{1,2}^{3}} & \frac{1}{l_{1,2}}
    \end{bmatrix} \\
    &\mathbf{H}_{\phi} = \frac{||{}^G\mathbf d||}{||{}^G\mathbf n||}{}^G\mathbf n+\frac{||{}^G\mathbf n||}{||{}^G\mathbf d||}\lfloor {}^G\mathbf p_C\rfloor {}^G\mathbf d 
    \end{aligned} 
\end{equation}
where $\mathbf J_L$ is the Jacobian of line projection, $l_{1,2}=\sqrt{l_1^2 + l_2^2}$.
\subsection{Vanishing Point Measurement Model} 

In the man-made environment, lines can be classified into structural lines with VP measurements and non-structural lines \cite{lim2022uv}. For a structural line, when its corresponding vanishing point is calculated, a VP measurement residual is obtained:
\begin{equation}
    \mathbf r_v = \mathbf{p}_v - \frac{1}{d_3}\begin{bmatrix}
    d_1 \\ d_2
    \end{bmatrix} \label{eq13}
\end{equation}
where $\mathbf r_v$ and $\mathbf p_v$ represent the vanishing point residual and the vanishing point observation, respectively. ${}^C\mathbf d = \begin{bmatrix}d_1 & d_2 & d_3\end{bmatrix}^T$ is the line direction vector in the camera frame, which is orthogonal
to the normal vector ${}^C\mathbf n$ which is used as the line measurement. In this sense, VP measurement is complementary to line measurement and fully exploits the line geometric information. The Jacobian with respect to the vanishing point is computed as follows:
\begin{equation}
\begin{aligned}
&\mathbf{H}_X = \mathbf{J}_L{}^G_C\mathbf R^T\begin{bmatrix} \lfloor {}^G\mathbf{d} \rfloor & \mathbf{0}_3 \end{bmatrix} \\
&\mathbf{H}_v = \mathbf{J}_L{}^G_C\mathbf R^T\begin{bmatrix} -\lfloor {}^G\mathbf{d} \rfloor & \frac{||{}^G\mathbf n||}{||{}^G\mathbf d||}{}^G\mathbf d \end{bmatrix}
\end{aligned}
\label{eq14}
\end{equation}
where the Jacobian items are similar to those in (\ref{eq29}).

\section{FEATURE INITIALIZATION}
\subsection{Line \& VP Detection}
When a new frame comes, line segments are detected by the LSD detector. For each line, the descriptor LBD is calculated for line tracking. Some outlier rejection strategies are employed such as the limited distance between the endpoints of matched lines and the limited angle difference between the line directions. As for VP detection, we use the efficient and robust two-line based VP detection algorithm \cite{lu20172}, and the observed lines with the detected VPs will be considered as the structural lines and used in the aforementioned line measurement model and VP measurement model.

\subsection{Line \& VP Estimation}
The strategy of line estimation is similar to that of point estimation in MSCKF. When a non-structural line is no longer tracked by the current camera frame, it will be triangulated by the dual Plucker matrix method \cite{he2018pl}. Then we optimize it by the following cost function:
\begin{equation}
    min \sum \limits_{i\in \mathcal{L}_n}||\mathbf{r}_L(\mathbf{o}_j,\mathbf l_i)||^2 \label{eq23}
\end{equation}
where $\mathcal{L}_n$ are the sets of non-structural line measurements in the sliding window, $\mathbf l_i$ is the $i$-th line measurement. After the structural line is triangulated, we add a VP cost item for a better estimation:
\begin{equation}
    min \sum \limits_{i\in \mathcal{L}_s}(||\mathbf{r}_L(\mathbf{o}_j,\mathbf l_i)||^2 + ||\mathbf{r}_v(\mathbf{o}_j, \mathbf p_{v,i})||^2) \label{eq24}
\end{equation}
where $\mathcal{L}_s$ are the sets of structural line measurement, $\mathbf p_{v,i}$ is the $i$-th vanishing point measurement.

\section{OBSERVABILITY ANALYSIS}
In this section, we perform the observability analysis of our proposed method based on the EKF model which has the same observability property as MSCKF. The observability matrix for EKF in the period [$t_m,t_n$] is defined as follows:
\begin{equation}
    \mathcal{O}_{n} \triangleq\begin{bmatrix}
    \mathbf{H}_{m} \\
    \mathbf{H}_{m+1}\Phi_{m} \\
    \vdots \\
    \mathbf{H}_{n} \Phi_{n-1} \cdots \Phi_{m}
    \end{bmatrix} \label{eq15}
\end{equation}
where $\mathbf H_k$ is the measurement Jacobian, $\mathbf \Phi_k$ is the state transition matrix from $t_k$ to $t_{k+1}$. Since previous literature has demonstrated that point-line based IEKF can improve the consistency without artificial remedies \cite{heo2018consistent}, we focus on the observability of the vanishing point. At time step $t_k$, the state vector is considered to contain only one line feature with the line and VP measurement. 
The observability matrix at $t_k$ can be written as:
\begin{equation}
    \mathcal{O}_{k}= \begin{bmatrix}
    \mathcal{O}_{l_{k}} \\
    \mathcal{O}_{v_{k}}
    \end{bmatrix} \label{eq16}
\end{equation}
The null space for $\mathcal{O}_{l_k}$ has been derived in \cite{heo2018consistent}, indicating that the unobservable subspace does not degenerate. And the second part can be derived as:
\begin{equation}
    \begin{aligned}
    \mathcal{O}_{v_{k}} &= \mathbf{J}_p {}^G_{I_k}\mathbf R^T\begin{bmatrix}
        \mathbf{\Gamma}_1 & \mathbf{\Gamma}_2
    \end{bmatrix} \\
    \mathbf{\Gamma}_1 &= \begin{bmatrix}\lfloor {}^G\mathbf{d} \rfloor & \mathbf{0}_{3\times 6} & \lfloor {}^G\mathbf{d} \rfloor \Phi_{14} & \mathbf{0}_3 \end{bmatrix} \\
    \mathbf{\Gamma}_2 &= \begin{bmatrix}-\lfloor {}^G\mathbf{d} \rfloor & \frac{||{}^G\mathbf n||}{||{}^G\mathbf d||}{}^G\mathbf d \end{bmatrix}
    \end{aligned}\label{eq17}
\end{equation}
There are at least eleven directions lying in the unobservable subspace with the null space matrix:
\begin{equation}
    \mathcal{N}_v = \begin{bmatrix}
        \mathbf{I}_3 & \mathbf{0}_{3} & \mathbf{0}_{3} & \mathbf 0_{3\times1} & \mathbf 0_{3\times1} \\
        \mathbf{0}_{3} & \mathbf{I}_{3} & \mathbf{0}_{3} & \mathbf 0_{3\times1} & \mathbf 0_{3\times1} \\
        \mathbf{0}_{3} & \mathbf{0}_{3} & \mathbf{I}_3 & \mathbf 0_{3\times1} & \mathbf 0_{3\times1} \\
        \mathbf{0}_{3} & \mathbf{0}_{3} & \mathbf{0}_{3} & \mathbf 0_{3\times1} & \mathbf 0_{3\times1} \\
        \mathbf{0}_{3} & \mathbf{0}_{3} & \mathbf{0}_{3} & \mathbf{0}_{3\times1} & \mathbf 0_{3\times1}\\
        \mathbf{I}_{3} & \mathbf{0}_{3} & \mathbf{0}_{3} & \mathbf {}^G \mathbf d & \mathbf 0_{3\times1}\\
        0 & 0 & 0 & 0 & ||{}^G\mathbf d||
    \end{bmatrix} \label{eq18}
\end{equation}
It is observed that the left nine columns of $\mathcal{N}_v$ are independent of the estimated state. Therefore, the IEKF model with line features can remain in the ideal unobservable subspace naturally. Meanwhile, there exist at least two unobservable directions for $\mathcal{O}_{lk}$:
\begin{equation}
    \mathcal{O}_{l_k}\mathcal{N}_{lk} = \mathcal{O}_{l_k}\begin{bmatrix}
        \mathbf 0_{15\times2}\\
        \mathcal{N}_{o_k}
    \end{bmatrix}=\mathbf 0 \label{eq30}
\end{equation}
where $\mathcal N_{l_k}$ is the null space matrix of $\mathcal{O}_{l_k}$, and $\mathcal{N}_{o_k}$ represents the null space regarding the line parameter. It can be verified that $\mathcal{O}_{v_k}\mathcal{N}_{l_k}\neq \mathbf 0$ which means the line is fully observable with VP measurements.
\begin{figure*}[t]
\subfigure[]{
    \centering
    \includegraphics[scale=0.4]{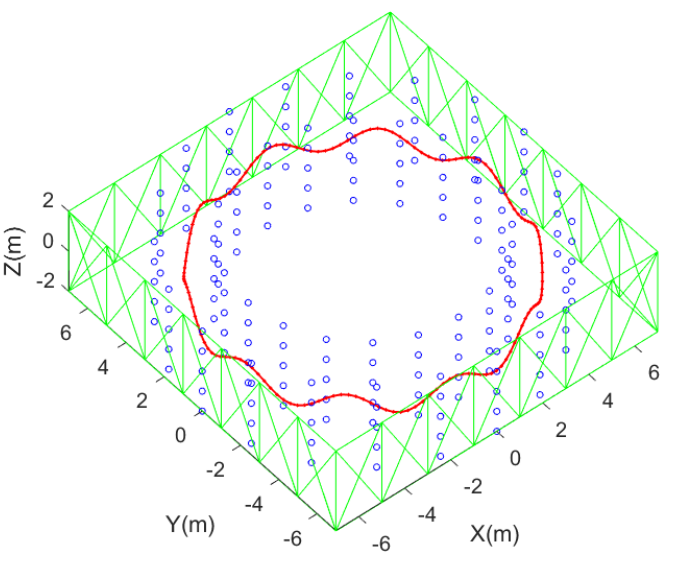}
}
\subfigure[]{
    \centering
    \includegraphics[scale=0.4]{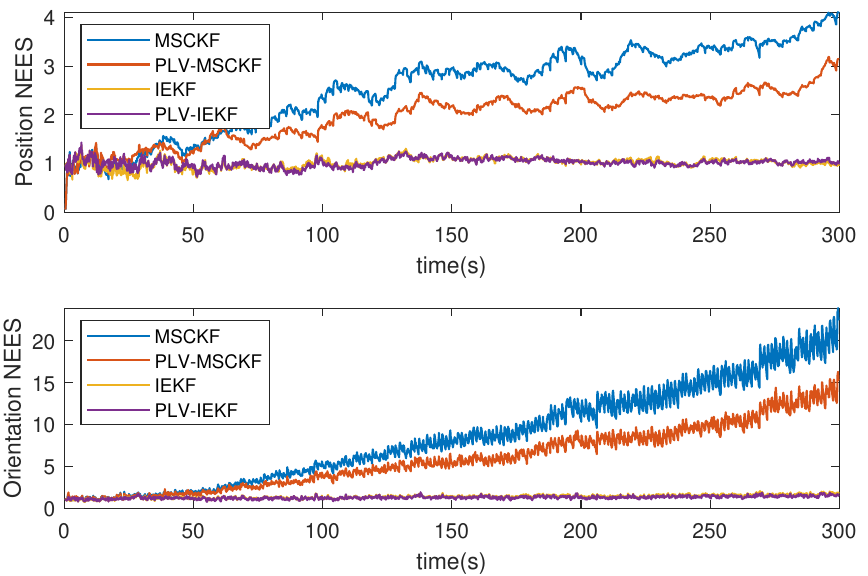}
}
\subfigure[]{
    \centering
    \includegraphics[scale=0.4]{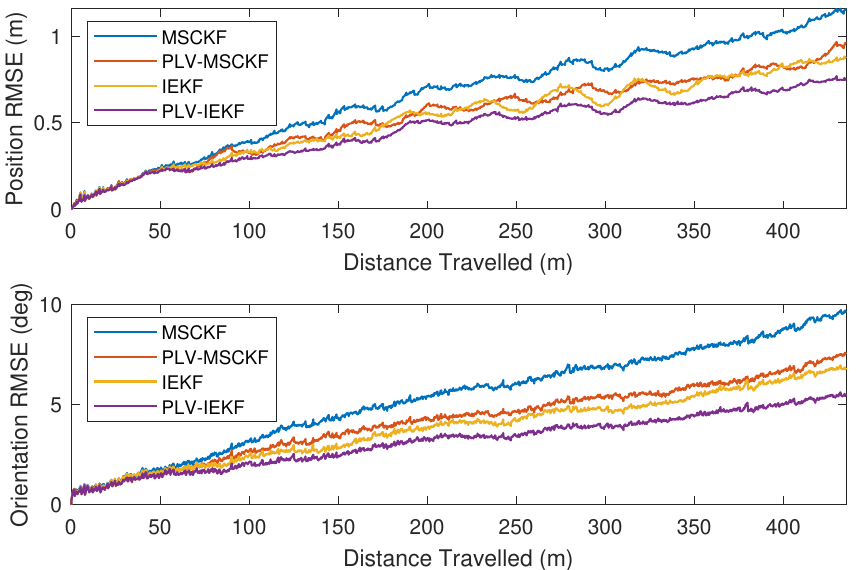}
}
\caption{Simulation trajectory and Monte Carlo results for 30 runs. (a): True trajectory (red), point landmarks (blue) and line landmarks (green); (b): Average NEES; (c): Average RMSE.}
\label{fig:rmse_nees}
\end{figure*}
\section{EVALUATION}
To validate the effectiveness of our proposed approach, we conduct the simulation and real-world test. In the real-world test, we compare our algorithm with some state-of-the-art VIO algorithms. Our experiments are conducted on a laptop with Intel(R) Core(TM) i7-10710U CPU@1.10Ghz and 16G RAM.
\subsection{Simulation}
In the simulation test, we assume that a car equipped with a 100Hz IMU and a 10Hz camera loops ten times along a circle with a radius of 6m. A total of 200 points and 140 lines are scattered around the real trajectory, as shown in Fig. \ref{fig:rmse_nees}(a). The points are generated along the inside cylinder wall with a radius of 5 meters and outside cylinder walls with a radius of 7 meters. And the lines are generated along a square wall with a length of 14 meters. The camera captures landmarks in the field of view of the camera within a range of 20 meters. We generate sensor measurements with noise according to the trajectory and landmarks. The initial noise matrix is set as $diag(0.008^2\mathbf{I}_3, 0.0004^2\mathbf{I}_3, 0.01^2\mathbf{I}_3, 0.003^2\mathbf{I}_3)$, and all the feature measurement noise is set as 1 pixel.

Four algorithms are compared in the simulation: standard point based MSCKF (MSCKF), standard point based IEKF, point-line-VP based MSCKF (PLV-MSCKF), and point-line-VP based IEKF (PLV-IEKF). The sliding window size is set as 20 and we only use the landmarks that are captured more than 5 times by the camera for robust estimation. The pose accuracy indicator Root Mean Squared Error (RMSE) and the consistency indicator Averaged Normalized Estimation Error Squared (ANEES) are reported in Fig. \ref{fig:rmse_nees}(b) and (c). It is evident that NEES of IEKF and PLV-IEKF is closer to the ideal NEES (one) than that of MSCKF family, indicating better consistency. Meanwhile, PLV-IEKF achieves a better pose estimation than IEKF with additional line and VP measurements.
\begin{figure}
    \centering
    \includegraphics[scale=0.6]{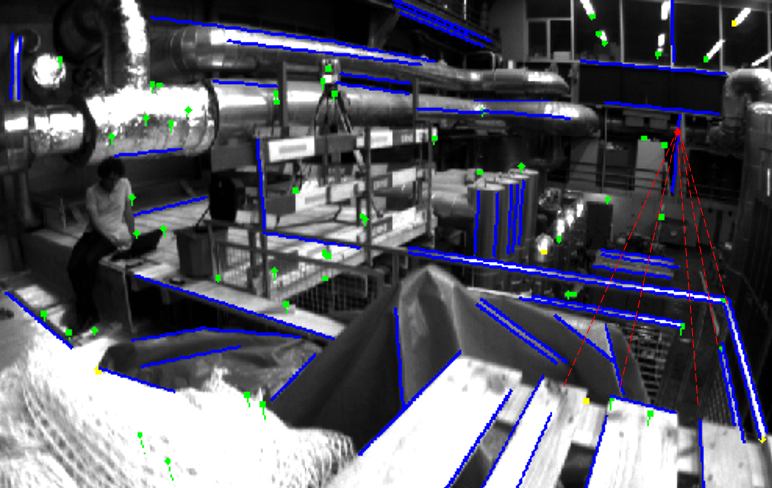}
    \caption{Tracked points (green), lines (blue), and vanishing points (red) in MH\_03\_medium sequence.}
    \label{fig:trackFeatures}
\end{figure}
\begin{table*}[t]
    \caption{Position RMSE(m) on the EuRoC dataset.}
    \label{tab2}
    \centering
    \renewcommand\arraystretch{1.5}
    \begin{tabular}{cccccccc}
        \hline
        \multirow{2}*{Sequence} & \multicolumn{7}{c}{Algorithm} \\
        \Xcline{2-8}{0.4pt}  & \makecell[c]{MSCKF} & \makecell[c]{PLV-MSCKF} & \makecell[c]{IEKF}  & \makecell[c]{PL-IEKF}  & \makecell[c]{VINS-Mono} & \makecell[c]{PL-VINS} & \makecell[c]{PLV-IEKF} \\
        \hline
        
        \makecell[c]{MH\_01\_easy} & \makecell[c]{0.265} & \makecell[c]{0.229}  & \makecell[c]{0.257}  & \makecell[c]{0.216}  & \makecell[c]{$\mathbf{0.154}$} & \makecell[c]{0.182} & \makecell[c]{\underline{0.174}}\\

        \makecell[c]{MH\_02\_easy} & \makecell[c]{0.295} & \makecell[c]{0.237}  & \makecell[c]{0.250}  & \makecell[c]{0.229}  & \makecell[c]{0.232} & \makecell[c]{\underline{0.208}} & \makecell[c]{$\mathbf{0.206}$}\\
        
        \makecell[c]{MH\_03\_medium} & \makecell[c]{0.443} & \makecell[c]{0.413}  & \makecell[c]{0.298}  & \makecell[c]{\underline{0.222}}  & \makecell[c]{0.269} & \makecell[c]{0.267} & \makecell[c]{$\mathbf{0.184}$}\\
        
        \makecell[c]{MH\_04\_difficult} & \makecell[c]{0.605} & \makecell[c]{\underline{0.346}}  & \makecell[c]{0.403}  & \makecell[c]{0.460}  & \makecell[c]{0.439} & \makecell[c]{0.409} & \makecell[c]{$\mathbf{0.333}$}\\
        
        \makecell[c]{MH\_05\_difficult} & \makecell[c]{0.404} & \makecell[c]{\underline{0.285}}  & \makecell[c]{0.294}  & \makecell[c]{0.325}  & \makecell[c]{0.300} & \makecell[c]{0.328} & \makecell[c]{$\mathbf{0.240}$}\\
        
        \makecell[c]{V1\_01\_easy} & \makecell[c]{0.105} & \makecell[c]{$\mathbf{0.070}$}  & \makecell[c]{0.098}  & \makecell[c]{0.079}  & \makecell[c]{0.082} & \makecell[c]{0.087} & \makecell[c]{\underline{0.073}}\\

        \makecell[c]{V1\_02\_medium} & \makecell[c]{0.144} & \makecell[c]{0.160}  & \makecell[c]{\underline{0.140}}  & \makecell[c]{0.164}  & \makecell[c]{$\mathbf{0.111}$} & \makecell[c]{0.171} & \makecell[c]{0.182}\\

        \makecell[c]{V1\_03\_difficult} & \makecell[c]{0.376} & \makecell[c]{0.308}  & \makecell[c]{0.331}  & \makecell[c]{0.272}  & \makecell[c]{0.250} & \makecell[c]{$\mathbf{0.148}$} & \makecell[c]{\underline{0.249}}\\

        \makecell[c]{V2\_01\_easy} & \makecell[c]{0.349} & \makecell[c]{0.154}  & \makecell[c]{0.242}  & \makecell[c]{0.187}  & \makecell[c]{0.165} & \makecell[c]{$\mathbf{0.126}$} & \makecell[c]{\underline{0.141}}\\

        \makecell[c]{V2\_02\_medium} & \makecell[c]{0.253} & \makecell[c]{0.240}  & \makecell[c]{0.245}  & \makecell[c]{0.268}  & \makecell[c]{\underline{0.209}} & \makecell[c]{$\mathbf{0.152}$} & \makecell[c]{0.224}\\

        \hline
        \multicolumn{8}{l}{* Bold and underline represent best and second best in each sequence, respectively.} 
    \end{tabular}
\end{table*}
\subsection{Real-world Test}
\begin{figure}
    \centering
    \includegraphics[scale = 0.6]{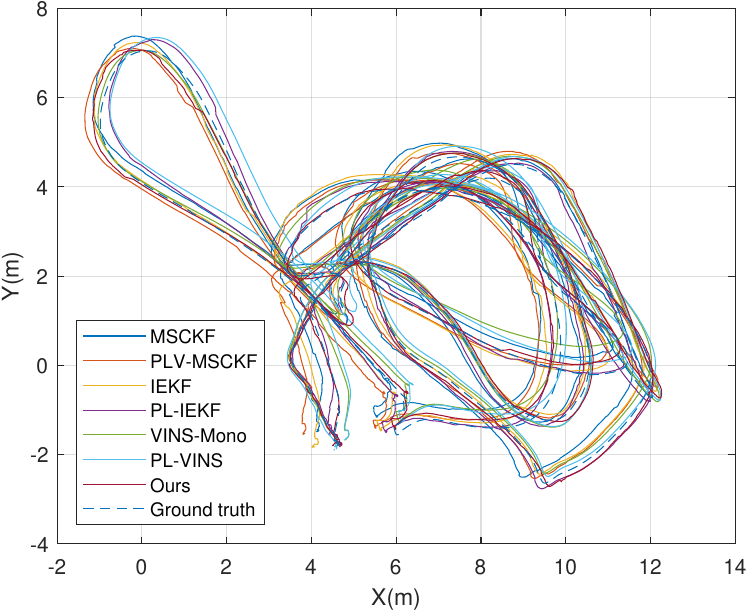}
    \caption{X-Y plot of estimated trajectories on the MH\_03\_medium sequence.}
    \label{fig:traj_euroc}
\end{figure}
To validate the effect of our proposed method, we conduct a real-world experiment on EuRoC Dataset \cite{burri2016euroc}, which exhibits varying levels of motion blur and textureless regions \cite{yang2019visual}. V2\_03\_difficult sequence is not chosen due to a number of missing camera frames which affect the feature tracking. We compare our algorithm with other state-of-the-art algorithms such as MSCKF \cite{sun2018robust}, IEKF \cite{brossard2018unscented}, VINS-Mono \cite{qin2018vins} and PL-VINS \cite{fu2020pl}. PLV-MSCKF and point-line based invariant IEKF (PL-IEKF) are also implemented for a detailed comparison. For all of the algorithms, we extract no less than 100 point features and 30 line features for robust estimation and disable loop detection for a fair comparison. The sliding window size for these filter-based VIO is set as 20. An example of tracked features in MH\_03\_medium is shown in Fig. \ref{fig:trackFeatures}. RMSE results are reported in Table \ref{tab2}, and the trajectory sample is shown in Fig. \ref{fig:traj_euroc}. 
It is evident that in most cases using lines and vanishing points can achieve a better pose accuracy, and PLV-IEKF has a better performance than PLV-MSCKF because of better consistency. 

It is also observed in Table \ref{tab2} that PLV-IEKF performs better in Machine Hall (MH) sequences than PL-IEKF since there exist more structural features in the environment, where vanishing point measurements can improve the line estimation, even in challenging sequences like MH\_04\_difficult sequence with illumination changing and textureless scenarios.
\begin{table*}[t]
    \caption{Backend processing time per frame on the MH\_01\_easy sequence.}
    \label{tab3}
    \centering
    \renewcommand\arraystretch{1.5}
    \begin{tabular}{cccccccc}
        \hline
        \makecell[c]{Algorithm} & \makecell[c]{MSCKF} & \makecell[c]{PLV-MSCKF} & \makecell[c]{IEKF}  & \makecell[c]{PL-IEKF}  & \makecell[c]{VINS-Mono} & \makecell[c]{PL-VINS} & \makecell[c]{PLV-IEKF}\\
        \hline
        \makecell[c]{Time cost (ms)} & \makecell[c]{5.21} & \makecell[c]{5.578} & \makecell[c]{5.33}  & \makecell[c]{5.43}  & \makecell[c]{37.9} & \makecell[c]{60.06} & \makecell[c]{5.65}\\
        \hline
    \end{tabular}
\end{table*}

Table \ref{tab3} shows the backend processing time on MH\_01\_easy sequence for different algorithms, among which the proposed PLV-IEKF requires slightly more time than other filter-based point-based and point-line-based methods, but much less time than VINS-Mono and PL-VINS. It demonstrates that PL-IEKF can achieve competitive and higher time efficiency compared with optimization-based methods.
\section{CONCLUSIONS}

In this paper, we propose an invariant Visual-Inertial Odometry (VIO) leveraging points, lines, and vanishing point features in man-made environments. Three feature measurement models are obtained in our framework, and two equivalent measurement models of points and lines are proven in our filter design. The observability matrix regarding vanishing points is derived to demonstrate that the EKF model with line features can guarantee the ideal unobservable subspace. Simulation tests and real-world experiments validate our algorithm's consistency and competitive accuracy compared with other state-of-the-art algorithms. In the future, we will refine the line and VP estimation to achieve an accurate and efficient mapping. 

\addtolength{\textheight}{-1cm}   



\section*{APPENDIX}

\subsection{Proof of Proposition 1}

To prove Proposition 1, we assume that only one point-type landmark in the current sliding window is observed by $n$ camera poses. When the landmark error is additive, the Jacobian for VIO state is given by:
\begin{equation}
\begin{aligned}
    \mathbf H_{x} &=
    \mathbf J_\pi{}\begin{bmatrix}
    \mathbf 0_{3\times15} & \mathbf H_{11} & \mathbf 0_{3\times6}  & ... &\mathbf 0_{3\times6}  \\
    \mathbf 0_{3\times15} & \mathbf 0_{3\times6} & \mathbf H_{22} & ... & \mathbf 0_{3\times6} \\
    ... & ... & ... & ... & ... \\
    \mathbf 0_{3\times15} & \mathbf 0_{3\times6} &  ...  & ... & \mathbf H_{nn}
    \end{bmatrix} \\
    \mathbf H_{ii} &= \begin{bmatrix}{}^{G}_{C_{i}}\mathbf R^T\lfloor {}^G \mathbf p_{f\times}\rfloor & -{}^{G}_{C_{i}}\mathbf R^T\end{bmatrix}, i = 1,2,...,n
\end{aligned}
\end{equation}
where $\mathbf J_\pi{} = \begin{bmatrix}
    \mathbf J_{p1} & \mathbf J_{p2} & \cdots & \mathbf J_{pn}
\end{bmatrix}$, $\mathbf J_{pi}$ is the Jacobian $\mathbf J_p$ for the $i$-th camera pose. On the other hand, the Jacobian regarding an invariant form is calculated as follows:
\begin{equation}
\begin{aligned}
\mathbf H'_{x} &= \mathbf J_\pi{}\begin{bmatrix}
\mathbf 0_{3\times15} & \mathbf H'_{11} & \mathbf 0_{3\times6} & ... &\mathbf 0_{3\times 6} \\
\mathbf 0_{3\times15} & \mathbf H'_{21} & \mathbf H'_{22} & ... &\mathbf 0_{3\times6} \\
... & ... & ... & ... & ...  \\
\mathbf 0_{3\times15} & \mathbf H'_{n1} &  ... & ... & \mathbf H'_{nn}
\end{bmatrix} \\
\mathbf H'_{11} &= \begin{bmatrix}
    \mathbf 0_3 & -{}^{G}_{C_{1}}\mathbf R^T
\end{bmatrix} \\
\mathbf H'_{i1} &= \begin{bmatrix}
-{}^{G}_{C_{i}}\mathbf R^T\lfloor {}^G \mathbf p_{f\times}\rfloor & \mathbf 0_3
\end{bmatrix}, \mathbf H'_{ii} = \mathbf H_{ii}, i = 2,3,4...,n
\end{aligned} \label{eq:appendix}
\end{equation}
Note that the Jacobians with respect to the landmark for both forms are the same:
\begin{equation}
\mathbf H_f = \mathbf J_\pi\begin{bmatrix}
{}^{G}_{C_{1}}\mathbf R & {}^{G}_{C_{2}}\mathbf R & ... & {}^{G}_{C_{n}}\mathbf R
\end{bmatrix}^T
\end{equation}
which shares the same left null space matrix $\mathbf{A}$. We calculate the difference between the final Jacobians:
\begin{equation}
\begin{aligned}
    &\mathbf A^T\mathbf H_x-\mathbf A^T\mathbf H'_x \\
    =&\begin{bmatrix}
        \mathbf 0_{3n\times 15} & \mathbf A^T\mathbf{H}_f\lfloor {}^G \mathbf p_{f\times}\rfloor & \mathbf 0_{3n\times (6n-3)}
    \end{bmatrix}\\
    =& \mathbf 0
\end{aligned}
\end{equation}
Therefore the final Jacobians are equivalent, so is the final projected measurement noise $\mathbf n_o = \mathbf A^T \mathbf n$.





\bibliographystyle{IEEEtran}
\bibliography{root}

\end{document}